\documentclass[conference]{ieeeconf}
\IEEEoverridecommandlockouts
\usepackage{graphicx}
\usepackage{cite}
\usepackage{picinpar}
\usepackage{amsmath}
\usepackage{url}
\usepackage[latin1]{inputenc}
\usepackage{colortbl}
\usepackage{soul}
\usepackage{multirow}
\usepackage{pifont}
\usepackage{color}
\usepackage{alltt}
\usepackage{enumerate}
\usepackage{siunitx}
\usepackage{epstopdf}
\usepackage{pbox}
\usepackage{amssymb}
\usepackage{mathtools}
\usepackage{bbm}
\usepackage{cite}
\usepackage{amsfonts}
\usepackage{algorithmic}
\usepackage{textcomp}
\usepackage{xcolor}

\usepackage{caption}

\definecolor{limegreen}{rgb}{0.2, 0.8, 0.2}
\definecolor{forestgreen}{rgb}{0.13, 0.55, 0.13}
\definecolor{greenhtml}{rgb}{0.0, 0.5, 0.0}
\definecolor{black}{rgb}{0.0, 0.0, 0.0}

\usepackage[colorlinks,allcolors=black]{hyperref}

\usepackage[font=small,labelfont=bf,tableposition=top]{caption}

\usepackage{blindtext}
\title{here title}

\let\oldtwocolumn\twocolumn
\renewcommand\twocolumn[1][]{%
    \oldtwocolumn[{#1}{
    \begin{center}
           \includegraphics[width=17.5cm]{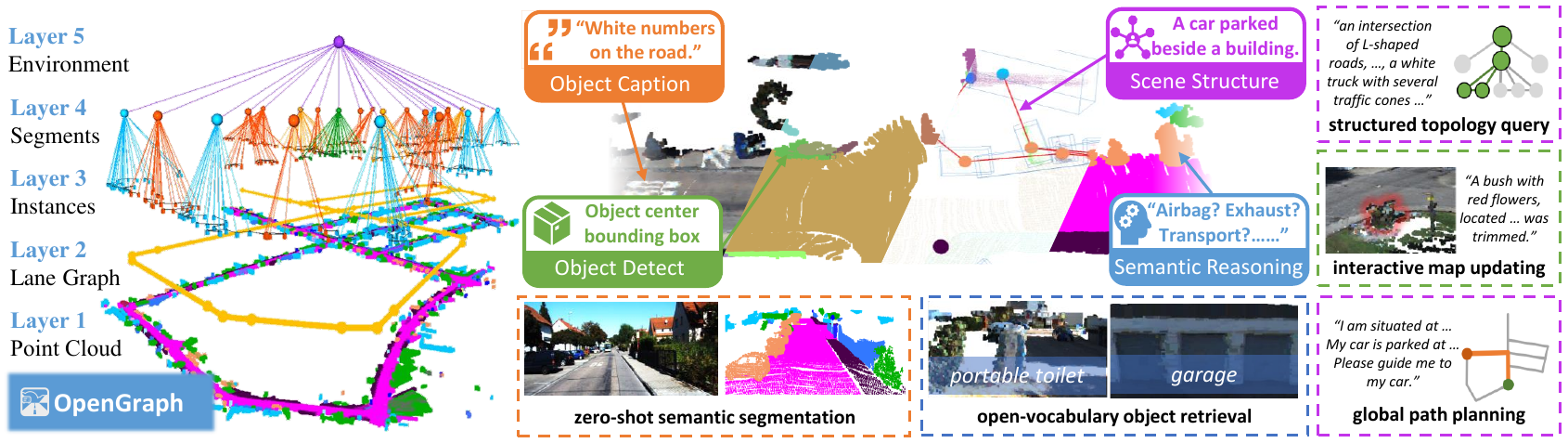}
           \captionof{figure}{We introduce \textbf{OpenGraph}, a framework of open-vocabulary hierarchical 3D graph representation in large-scale outdoor environments. OpenGraph facilitates various downstream tasks, including zero-shot semantic segmentation, open-vocabulary object retrieval, structured topology query, global path planning, interactive map updating, and so on.}
           \label{first}
        \end{center}
    }]
}

\begin{document}

\pagestyle{plain}
\title
{	
\textbf{OpenGraph: Open-Vocabulary Hierarchical 3D Graph Representation in Large-Scale Outdoor Environments}
    
}

\author{
Yinan Deng, Jiahui Wang, Jingyu Zhao, Xinyu Tian, Guangyan Chen, Yi Yang, Yufeng Yue$^{*}$
\thanks{This work is supported by the National Natural Science Foundation of China under Grant  62003039, 61973034, U193203, 62173042. (Corresponding Author: Yufeng Yue, yueyufeng@bit.edu.cn)}
\thanks{All authors are with School of Automation, Beijing Institute of Technology, Beijing, 100081, China.}
}

\maketitle

\begin{abstract}

Environment representations endowed with sophisticated semantics are pivotal for facilitating seamless interaction between robots and humans, enabling them to effectively carry out various tasks. Open-vocabulary maps, powered by Visual-Language models (VLMs), possess inherent advantages, including zero-shot learning and support for open-set classes. However, existing open-vocabulary maps are primarily designed for small-scale environments, such as desktops or rooms, and are typically geared towards limited-area tasks involving robotic indoor navigation or in-place manipulation. They face challenges in direct generalization to outdoor environments characterized by numerous objects and complex tasks, owing to limitations in both understanding level and map structure. In this work, we propose OpenGraph, the first open-vocabulary hierarchical graph representation designed for large-scale outdoor environments.  OpenGraph initially extracts instances and their captions from visual images, enhancing textual reasoning by encoding them. Subsequently, it achieves 3D incremental object-centric mapping with feature embedding by projecting images onto LiDAR point clouds. Finally, the environment is segmented based on lane graph connectivity to construct a hierarchical graph. Validation results from public dataset SemanticKITTI demonstrate that OpenGraph achieves the highest segmentation and query accuracy. The source code of OpenGraph is publicly available at \urlstyle{tt} \underline{\url{https://github.com/BIT-DYN/OpenGraph}}. 

\end{abstract}

\section{Introduction}

A comprehensive understanding and representation of the 3D scene are crucial for robots to perform various downstream tasks \cite{MACIM}. Occupancy mapping \cite{OctoMap} stands out as the most prevalent technique for building maps, allowing for the retrieval of geometric scene properties. By discerning obstacle positions and shapes, these maps facilitate spatial navigation for autonomous obstacle avoidance.
With the advancement of deep learning technology, semantic information is incorporated into classic geometric maps \cite{see-csom}. This integration enables robots to achieve semantic-level intelligent navigation. 
However, such semantics are confined to predefined labels during the training phase, presenting challenges in heuristic comprehension and effective utilization.

In recent years, the widespread adoption of visual language models (VLMs) \cite{CLIP, ALIGN,blip}  has opened avenues for encapsulating conceptual semantics in maps. VLMs encode images and text into a unified feature space through adversarial learning, enabling seamless interaction between robots and humans. These foundation models, trained on extensive web-based datasets, can uncover novel objects and derive simplistic understandings at inference time. 
However, the main scope of current open-vocabulary mapping methods is at room-level or desktop-level, primarily used for indoor navigation or in-place manipulation tasks for robots. There are two notable limitations, that restrict their applicability in large-scale outdoor environments:

1) \textbf{Weak object-centric comprehension and reasoning capabilities.} 
Most existing methods directly distil \cite{Clip-fields}  or project \cite{Conceptfusion} 2D VLM features into 3D space as semantic understanding of the constructed open-vocabulary maps. Such primitive VLM features excel at broad recognition but lack robust reasoning capabilities. 
For instance, when encountering \textit{a patch of grass}, VLM features capture its category (\textit{grass}), color (\textit{green}), and other basic attributes but not encompass common-sense knowledge such as  \textit{its function as a soccer field or a primary food for sheep}. 
This limited understanding restricts their applicability when handling various tasks encountered in outdoor environments that demand a certain level of comprehension.

2) \textbf{Limited and inefficient map architectures.}  
Efficiently storing, maintaining, and rapidly retrieving desired objects in non-structured outdoor environments characterized by numerous objects pose a key challenge.
The original point-wise open-vocabulary mapping methods \cite{Conceptfusion,lerf} are computationally expensive and typically represent retrieval results using feature similarity heatmaps, lacking object boundaries. Although some subsequent methods \cite{ovir, openmask3d}  have achieved indoor instance-level map construction with the assistance of segmentation models like SAM \cite{sam}, they remain challenging to effectively discriminate the desired one when multiple corresponding objects are present in the scene.  This difficulty is compounded by the prevalence of repetitive objects in outdoor settings, making it impractical to directly apply existing open-vocabulary map architectures to such environments.

To address the above limitations, this paper proposes \textbf{OpenGraph}, a novel framework for open-vocabulary hierarchical 3D graph representation.
\textbf{1) \underline{Open} reasoning:} Unlike approaches that directly employ VLM features for environmental semantic understanding, OpenGraph leverages VLMs as the cognitive front-end. It segments instances from visual images and generates textual captions. Moreover, large language models (LLMs) \cite{Llama, sbert}, renowned for their exceptional performance in natural language processing tasks, encode these captions to enrich the open-minded reasoning capabilities.
\textbf{2) Hierarchical \underline{graph}:} OpenGraph projects caption features from 2D images onto 3D LiDAR point clouds and incorporates them into the construction of object-centric maps. To facilitate map maintenance and specific object retrieval, OpenGraph introduces a hierarchical graph representation. This representation segments the environment based on the connectivity of the computed lane map and associates it with jurisdictional instances.
An illustrative example of OpenGraph's result is presented in Fig. \ref{first}. In summary, our contributions are as follows:

\begin{itemize}
    \item We introduce the first outdoor open-vocabulary object-centric mapping system capable of discovering, building, and comprehending a vast number of instances.
    We innovatively design the caption feature as the cornerstone for object comprehension, thereby enhancing the cognitive level of the maps.
    \item We propose a hierarchical 3D graphical representation that supports efficient maintenance and rapid retrieval in large-scale environments.
    \item Validation on outdoor dataset demonstrates that OpenGraph enables a profound semantic understanding of the environment and facilitates downstream applications.
\end{itemize}


\begin{figure*}[!t]\centering
	\includegraphics[width=17.5cm]{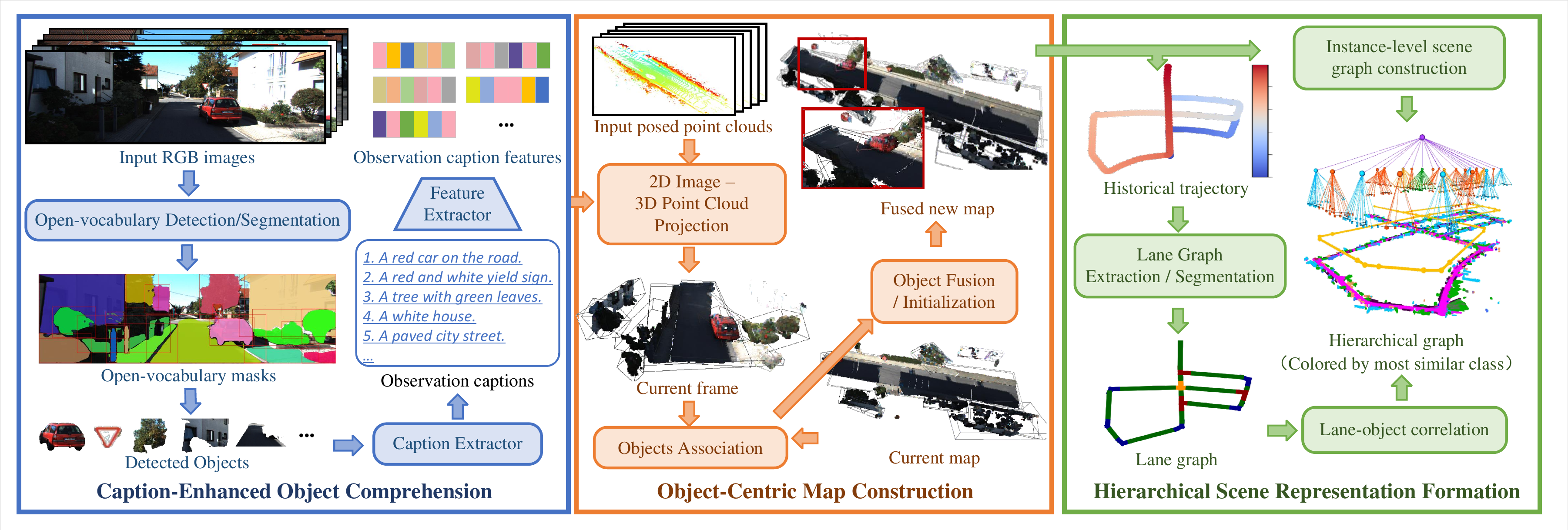}
	\caption{The framework of OpenGraph consists of three main modules: Caption-Enhanced Object Comprehension, Object-Centric Map Construction, and Hierarchical Graph Representation Formation. } 
	\label{framework} 
\end{figure*}

\section{Related Works} \label{RW}

\subsection{Closed-vocabulary semantic mapping}
 While early studies \cite{yang2017semantic, S-MKI, hd-ccsom}  introduced pretrained deep learning-based segmentation models into basic spatial representations (occupancies, point clouds, etc.) for semantic 3D mapping, their performance was largely constrained by the model capabilities. Recent researches \cite{shi2024city} exploit the latest advances in implicit neural representations to achieve geometric and semantic mapping within a unified feature space. Despite these advances, these methods require time-consuming self-supervision and are scenario-tailored, making generalization to other scenarios challenging. Most importantly, all of them either use segmentation models pretrained on a closed set of classes or can only utilize limited classes in the current scene for learning from scratch. This limitation makes them challenging to comprehend unseen object classes in complicated and open scenes. 
 
\subsection{Open-vocabulary 3D mapping}
With the impressive progress of web-based pre-trained visual language models and large language models, an increasing number of methods are attempting to extend their 2D open vocabulary understanding capabilities to the 3D world, including the following three mainstream solutions.

1) \textbf{Vision-only point-wise mapping}. Originally, OpenScene\cite{openscene} achieves 3D open-vocabulary scene understanding by projecting point-wise vision features extracted from fine-tuned image segmentation model \cite{Lseg} onto a 3D point cloud, facilitating easy utilization for open-vocabulary queries. However, the fine-tuned models lose their original ability to capture long-tail objects. Even if subsequent works \cite{Conceptfusion,lerf} adopt well-designed point-wise feature extraction methods to avoid object forgetting, the inconsistency between point features blurs the boundaries of instances, making object retrieval challenging during downstream robot tasks. 

2) \textbf{Vision-only instance-level mapping}. The mainstream method for instance-level open-vocabulary 3D mapping involves extracting VLM features at the 2D image mask scale and then fusing point clouds from potentially the same instances in 3D space, considering both spatial and feature similarities. Compared to OVIR-3D \cite{ovir}, which fuses text-aligned 2D region proposals into 3D space using Periodic 3D Instance Filtering and Merging, OpenMask3d \cite{openmask3d} leverages predicted class-agnostic 3D instance masks to guide the multi-view fusion of CLIP-based image embeddings. Recently, Open3DIS \cite{open3dis} devises a 2D-guide-3D Instance Proposal Module to further enhance the description of 3D object shapes. However, the above methods still rely on fused CLIP features, which encode limited visual context under multi-view masks, thus lacking high-level natural language reasoning capabilities. 

3) \textbf{Vision-language mapping}. Although not instance-level, the weakly supervised CLIP-Fields \cite{Clip-fields} first combines visual features from CLIP and textual features from Sentence-BERT \cite{sbert} in the architecture of neural implicit representations as robotic semantic memory. Beyond \cite{ovir}, ConceptGraphs \cite{conceptgraphs} further extracts structural captions for each object, feeding them to LLMs for other applications, such as LLM-powered scene graph creation and natural language reasoning. Nonetheless, their reasoning capabilities are still constrained by the utilization of underlying CLIP features, and their effectiveness in large-scale outdoor scenes is impeded by the reliance on depth cameras.

For OpenGraph, we employ the advanced vision language model to extract visual-textural captions directly. These captions are then encoded using LLM, yielding textual features enriched with natural language reasoning abilities. 

\subsection{3D Scene graph}
To overcome ambiguity in object retrieval by providing context-aware specifications, 3D scene graphs (3DSGs) are proposed to describe the 3D scene compactly as graph structures, where nodes represent spatial or semantic properties of objects and edges encode inter-object relationships \cite{hydra}. While original approaches \cite{scenegraphfusion} generate real-time closed-set 3D semantic scene graph predictions from image sequences using graph neural networks, recent researchers have explored integrating open-vocabulary foundation models for 3D scene graph generation. ConceptGraphs \cite{conceptgraphs} and OVSG \cite{OVSG} pioneered an open-vocabulary framework for generating indoor robot scene graphs. However, their performance in outdoor environments is compromised by focusing solely on establishing object-level scene graphs. Inspired by closed-vocabulary 3D hierarchical graph generation methods hydra \cite{hydra} and CURB-SG \cite{greve2023collaborative}, our OpenGraph proposes the first open-vocabulary hierarchical 3D graph representation in large-scale outdoor environments.

\section{OpenGraph}  \label{method}

\subsection{Framework Overview}

OpenGraph takes a sequence of 2D RGB images $\mathcal{I}=\{I^{(1)},I^{(2)},...,I^{(t)}\}$ and a sequence of 3D LiDAR point clouds $\mathcal{C}=\{C^{(1)},C^{(2)},...,C^{(t)}\}$ with pose $\mathcal{P}=\{P^{(1)},P^{(2)},...,P^{(t)}\}$ as input, and produces a global hierarchical graph $\mathcal{M}_{all}^{(t)}$ of the observed environment as output.

The overall OpenGraph framework, as depicted in Fig. \ref{framework}, consists of three primary modules. 
Firstly, the Caption-Enhanced Object Comprehension module focuses on instance segmentation and caption feature extraction from 2D images. Secondly, the Object-Centric Map Construction module is responsible for projecting 2D images and their interpretations into 3D LiDAR point clouds, enabling incremental construction of object-centric maps. Lastly, the Hierarchical Graph Representation Formation module deals with lane graph extraction and segmentation to construct the final hierarchical graph. By sequentially executing the three modules, OpenGraph acquires a profound understanding of the environment. 
The resulting hierarchical graph consists of the following layers, which can be expanded or collapsed as required by the actual scenario:
\begin{enumerate}
\item \textbf{\textit{Point Cloud Layer}} represents the metric point cloud $\mathcal{M}_{pc}$, providing the most intuitive representation of the environment.
\item \textbf{\textit{Lane Graph Layer}} depicts the lane graph $\mathcal{M}_{lg}$, which contains its own topology and can be utilized for downstream tasks such as path planning.
\item \textbf{\textit{Instance Layer}} is a subgraph of instances $\mathcal{M}_{ins}=\langle \textbf{O}, \textbf{E} \rangle$, where instances $\textbf{O}$ are composed of their centers of mass, bounding boxes, captions, and high-dimensional semantic features. Spatial relationships between instances are represented by edges $\textbf{E}$ connecting them.
\item \textbf{\textit{Segment Layer}} represents segments $\mathcal{M}_{seg}$, partitioned based on the connectivity of lane graph $\mathcal{M}_{lg}$, with each one having a center of mass.
\item \textbf{\textit{Environment Layer}} encompasses the entire outdoor environment node $\mathcal{M}_{env}$ connected to all segments.
\end{enumerate}
Edges connect nodes within each layer (e.g., to model traversability between segments) or across layers (e.g., to model that point clouds belong to an instance, or that an instance is in a certain segments).

\begin{figure}[!t]\centering
	\includegraphics[width=8cm]{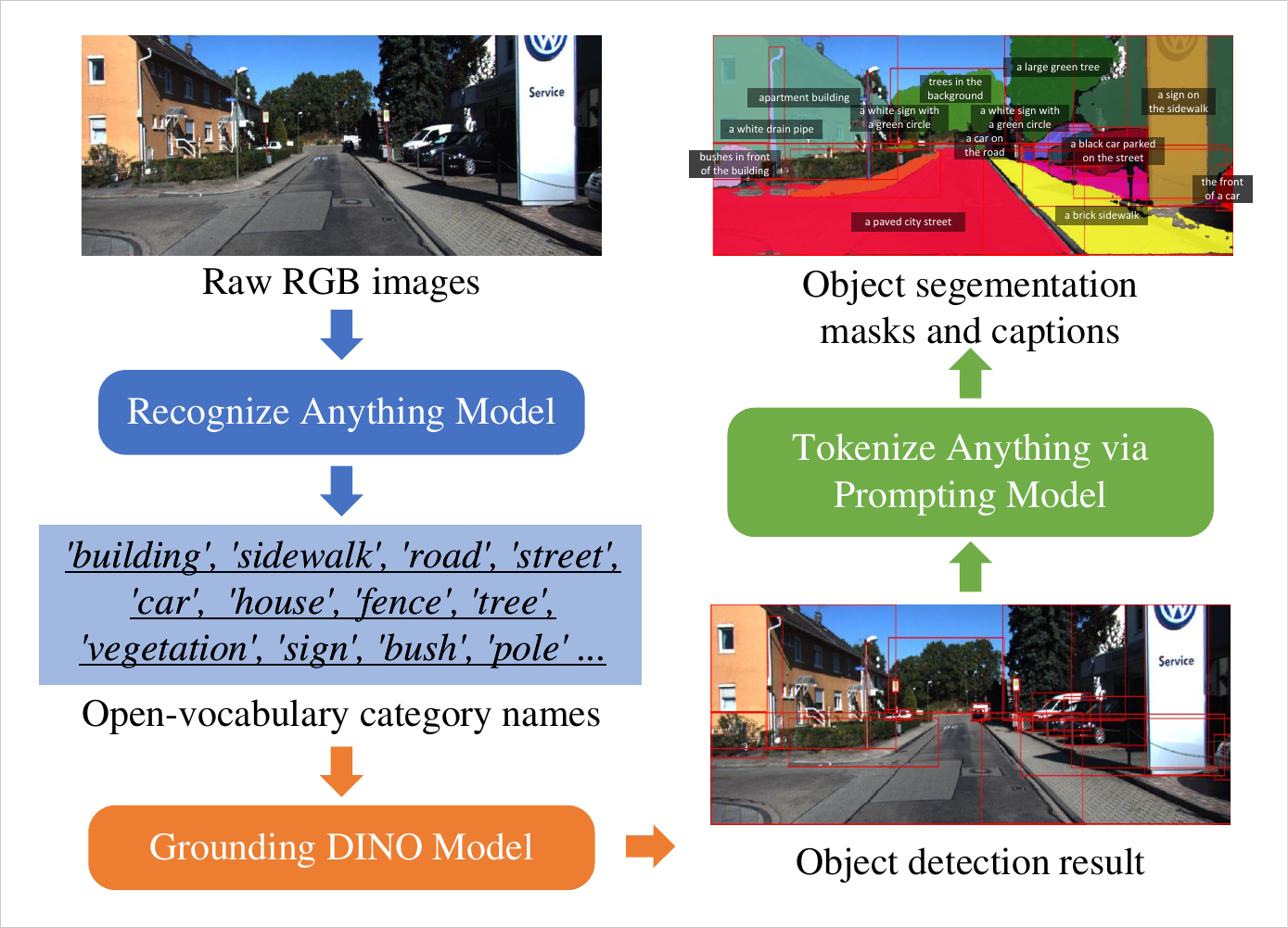}
	\caption{We employ three sequential visual language models for image instance segmentation and caption extraction. These models sequentially perform recognition, detection, simultaneous segmentation, and description generation of objects within the input image.} 
	\label{caption} 
\end{figure}

\subsection{Caption-Enhanced Object Comprehension}
Vision provides a wealth of information about the shape, size, texture, and location of objects and is an important perception for understanding the environment. The Caption-Enhanced Object Comprehension module sequentially uses three visual language models to realize the caption extraction of objects in each image frame as shown in Fig. \ref{caption}. 

At the current time $t$, for the input RGB image $I^{(t)}$, we first utilize Recognize Anything Model (RAM) \cite{ram} $\operatorname{Rec}(\cdot)$ to recognize the categories present in it. With its open-set capability, RAM is feasible to recognize any common category. Subsequently, we feed the generated open-vocabulary category names along with the original image into the Grounding DINO model \cite{groundingdino} $\operatorname{Det}(\cdot,\cdot)$ for open-set object detection. This yields the object detection bounding boxes, which are precisely what is required for the TAP (Tokenize Anything via Prompting) model \cite{tap} $\operatorname{SegCap}(\cdot,\cdot)$. Prompted by the object detection bounding boxes, the TAP model segments and describes the main objects within them, producing a set of masks $\{m^{(t)}_i\}_{i=1,,,m}$ and a set of captions $\{c^{(t)}_i\}_{i=1,,,m}$ for the current frame $I^{(t)}$. The entire process can be represented as (\ref{cap}), and the specific model can be substituted with others possessing similar functionality.
\begin{equation}
\begin{aligned}
    \label{cap}
    \{m_{i}^{(t)},c_{i}^{(t)}\}=\operatorname{SegCap}\left( {{I}^{(t)}},\operatorname{Det}\left( {{I}^{(t)}},\operatorname{Rec}({{I}^{(t)}}) \right) \right)
\end{aligned}
\end{equation}

VLMs aid in expressing the visual semantic understanding of objects through caption text, thereby facilitating the conversion from visual to language modality. However, the captions $\{c^{(t)}_i\}$ lack inherent common sense reasoning abilities. Conversely, LLMs find widespread application across various natural language processing tasks, having been pre-trained on extensive text datasets. Hence, as depicted in (\ref{feat}), we leverage LLMs to encode object captions, thereby generating high-dimensional embedded features to enhance comprehension and reasoning.   Consequently, we employ the SBERT model \cite{sbert} in our experiments.
\begin{equation}
\begin{aligned}
    \label{feat}
    \textbf{f}_{i}^{(t)}=\operatorname{Embed}(c_{i}^{(t)})
\end{aligned}
\end{equation}

After processing with the foundation model described above, we derive the 2D masks $\{m^{(t)}_i\}$ and captions $\{c^{(t)}_i\}$ for candidate objects, along with their corresponding caption features $\{\textbf{f}^{(t)}_i\}$, based on the input RGB images $I^{(t)}$ observed at the current time.

\subsection{Object-Centric Map Construction}

To achieve 3D mapping, it's crucial to incorporate metric information regarding detected objects. We employ a multi-sensor calibration and fusion method to project the 3D point cloud acquired from LiDAR onto a 2D image plane. Before projection, we employ 4DMOS \cite{4dmos} to detect and filter dynamic points from the point cloud $C^{(t)}$ to eliminate trailing shadows. The projection process generates 3D point clouds $\textbf{p}_{i}^{(t)}$ of the object instances that are aligned with the corresponding masks $m_{i}^{(t)}$:
\begin{equation}
\begin{aligned}
    \label{project}
    \textbf{p}_{i}^{(t)}=\{{{l}_{k}}|{K}T{{l}_{k}}\in m_{i}^{(t)}, {l}_{k}\in C^{(t)} \}
\end{aligned}
\end{equation}
where ${l}_{k}$ is the LiDAR point, $K$ is the internal camera parameter, and $T$ is the external lidar-to-camera parameter. Considering the existence of bias in the calibration parameters and the noise of the sensors themselves, we use the DBSCAN clustering algorithm to reduce the noise of the point clouds $\textbf{p}_{i}^{(t)}$ for each object and transform them to the map frame according to the pose ${P}^{(t)}$.

For objects $\textbf{o}^{(t)}_i=\langle \textbf{p}_{i}^{(t)}, c_{i}^{(t)}, \textbf{f}_{i}^{(t)} \rangle$ detected in the current frame, we need to fuse them into the existing map $\textbf{M}^{(t-1)} =\{\textbf{m}^{(t-1)}_j\}=\{\langle \textbf{p}_{j}^{(t-1)}, c_{j}^{(t-1)}, \textbf{f}_{j}^{(t-1)} \rangle\}$. We compute the geometric similarity ${{\phi}_{geo}}(i,j)$, caption similarity ${{\phi}_{cap}}(i,j)$ and feature similarity ${{\phi}_{fea}}(i,j)$ of each newly detected object with all objects in the map. Geometric similarity ${{\phi}_{geo}}(i,j)$ represents the 3D bounding box Intersection over Union (IoU) of point clouds $\textbf{p}_{i}^{(t)}$ and $\textbf{p}_{j}^{(t-1)}$. Caption similarity ${{\phi}_{cap}}(i,j)$ is calculated using cosine similarity after vectorizing the caption text $c_{i}^{(t)}$ and $c_{j}^{(t-1)}$ with TF-IDF (Term Frequency-Inverse Document Frequency). Feature similarity ${{\phi}_{fea}}(i,j)$ is the cosine similarity between the two features $\textbf{f}_{i}^{(t)}$ and $\textbf{f}_{j}^{(t-1)}$. The overall similarity measure $\phi (i,j)$ is obtained as the weighted sum of the three similarities:
\begin{equation}
\begin{aligned}
    \label{similarity}
    \phi (i,j)={{\omega }_{geo}}{{\phi }_{geo}}+{{\omega }_{cap}}{{\phi }_{cap}}+{{\omega }_{fea}}{{\phi }_{fea}}
\end{aligned}
\end{equation}

We perform object association using a greedy assignment strategy, where each newly detected object $\textbf{o}^{(t)}_i$ is paired with the existing object $\textbf{m}^{(t-1)}_j$ possessing the highest similarity score. If an object fails to find a match with a similarity score exceeding the threshold ${{\delta }_{sim}}$, it is initialized as a new object in the map $\textbf{M}^{(t)}$. For associated objects, we conduct object fusion. This involves merging the point clouds as $\textbf{p}_{j}^{(t)} = \textbf{p}_{i}^{(t)} \cup \textbf{p}_{j}^{(t-1)}$ and updating the features as 
\begin{equation}
\begin{aligned}
    \label{feat_up}
    {\textbf{f}}_{j}^{(t)}=({\textbf{f}}_{i}^{(t)}+n_{\textbf{m}_j}\textbf{f}_{j}^{(t)})/(n_{\textbf{o}_j}+1)
\end{aligned}
\end{equation}
where $n_{\textbf{m}_j}$ is the number of detections associated with $\textbf{m}_j$ so far. Additionally, to merge textual captions $c_{i}^{(t)}$ and $c_{j}^{(t-1)}$, we utilize the open-source LLM LLaMA \cite{Llama} to ensure a comprehensive, conflict-free description through prompts.

After incorporating all the objects $\textbf{o}^{(t)}_i$ of the current frame into the map, we get the incrementally updated map $\textbf{M}^{(t)}=\{\textbf{m}^{(t)}_j\}=\{\langle \textbf{p}_{j}^{(t)}, c_{j}^{(t)}, \textbf{f}_{j}^{(t)} \rangle\}$.

\begin{figure}[!t]\centering
	\includegraphics[width=6.5cm]{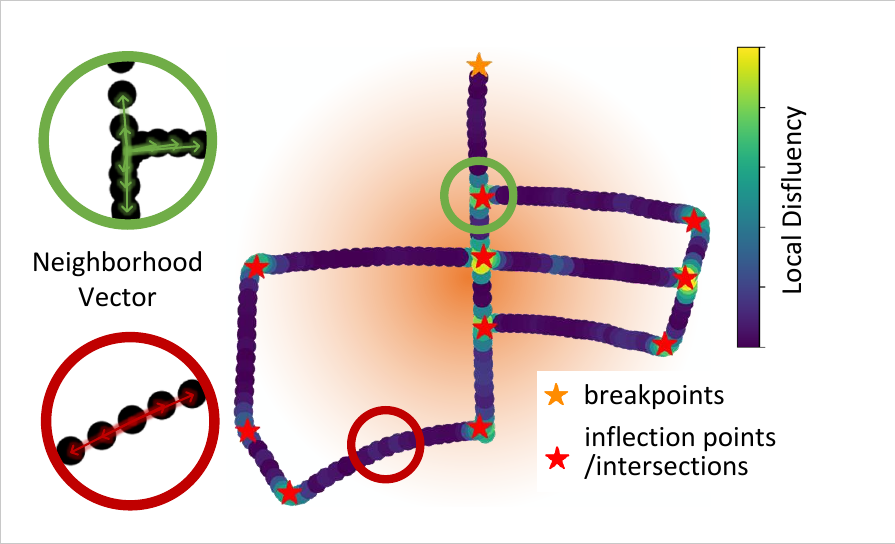}
	\caption{We extract the lane graph $\mathcal{M}_{lg}$ from historical trajectories $P^{(1:t)}$, whose nodes are derived from vector pinch angle ${{\Theta }^{(n)}}$ (breakpoints) and local disfluency ${{\lambda }^{(n)}}$ (inflection points or intersections).} 
\label{lane}    
\end{figure}

\begin{figure*}[!t]\centering
	\includegraphics[width=17.5cm]{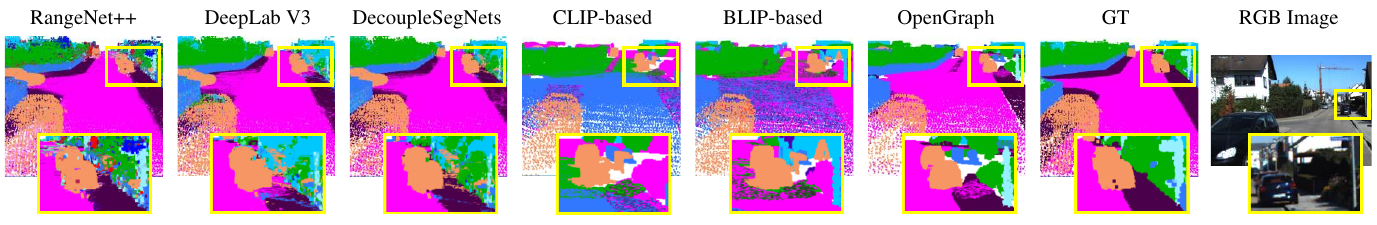}
	\caption{Semantic segmentation results on the SemanticKITTI dataset, utilizing 19 classes, indicate that despite not undergoing fine-tuning, OpenGraph demonstrates higher segmentation accuracy and reduced noise levels.} 
	\label{ss} 
\end{figure*}

\subsection{Hierarchical Graph Representation Formation} \label{hsgf}
On the basis of the map $\textbf{M}^{(t)}$, we generate the hierarchical graph representation $\mathcal{M}_{all}^{(t)}$. 

For \textbf{\textit{Point Cloud Layer}}, we stack the point clouds of objects as $\mathcal{M}_{pc}=\bigcup_j{\textbf{p}_{j}^{(t)}}$ and downsample them. To enhance the visualization, we can render the point cloud using the commonly used colors of the 19 outdoor classes.
Specifically, we assign a class to the point cloud  $\textbf{p}_{j}^{(t)}$ that has the highest similarity to the caption $\textbf{c}_{j}^{(t)}$ among all classes.

For \textbf{\textit{Lane Graph Layer}},  lane graph $\mathcal{M}_{lg}$ is derived from historical trajectories $P^{(1:t)}$. Initially, trajectories $P^{(1:t)}$ are projected onto a 2D plane as $\widetilde{P}^{(1:t)}$ by eliminating vertical displacements. Nodes in the lane graph $\mathcal{M}_{lg}$ represent inflection points, intersections, and breakpoints. To detect them, we assess local disfluency $\lambda^{(n)}$ at each trajectory point $\widetilde{P}^{(n)}$. Specifically, we consider a neighborhood within a radius $R$, forming sets of neighborhood vectors as
\begin{subequations}
\begin{align}
    \label{vectors}
    \textbf{V}^{(n)} &= \mathcal{N}\left( {{\widetilde{P}}^{(n)}} \right)-{{\widetilde{P}}^{(n)}}\\ 
    \label{Np}
    \mathcal{N}\left( {{\widetilde{P}}^{(n)}} \right)&=\left\{ {{\widetilde{P}}^{(m)}}\left| \left| {{\widetilde{P}}^{(m)}}-{{\widetilde{P}}^{(n)}} \right|<R \right. \right\}
\end{align} 
\end{subequations}
The angles are computed pairwise within the set of vectors, and the difference between each angle and $0$ or $\pi$ is calculated. Then, the average of the smaller values is determined and considered as the local disfluency measure:
\begin{subequations}
\begin{align}
    \label{theta}
    {{\Theta }^{(n)}}&=\left\{ \arccos \left( \frac{{{v}_{i}}\cdot {{v}_{j}}}{\left\| {{v}_{i}} \right\|\cdot \left\| {{v}_{j}} \right\|} \right) \right\}, \forall {{v}_{i}},{{v}_{j}}\in {\textbf{V}^{(n)}}  \\
    \label{lambda}
    {{\lambda }^{(n)}}&=\operatorname{mean}\left( \min \left( \left| {{\theta }_{i}} \right|,\left| {{\theta }_{i}}-\pi  \right| \right) \right), \forall  {{\theta }_{i}}\in {{\Theta }^{(n)}}
\end{align} 
\end{subequations}

We employed DBSCAN to cluster trajectory points exhibiting disfluency exceeding a threshold ${{\delta }_{dis}}$, thereby identifying inflection points and intersections. To discern breakpoints, we assess whether the mean of ${{\Theta }^{(n)}}$ is proximate to $0$. Fig. \ref{lane} exemplifies the detection of lane graph nodes. The edges of the lane graph $\mathcal{M}_{lg}$ are subsequently derived from the inter-node links and are trimmed based on trajectory alignment.

For \textbf{\textit{Instance Layer}}, we calculate the bounding box and center of mass for each object $\textbf{m}^{(t)}_j$ within the object-centric map $\textbf{M}^{(t)}$, treating them as nodes $\textbf{O}$.  Subsequently, we compute the bounding box IoUs between pairs of objects, resulting in a dense graph which we refine by estimating a Minimum Spanning Tree (MST). Additionally, to infer relations among nodes, we feed the captions, poses and bounding boxes of object pairs into LLaMA, yielding an open-vocabulary instance-layer scene graph $\mathcal{M}_{ins}=\langle \textbf{O}, \textbf{E} \rangle$. The nodes $\textbf{O}$ are connected to the corresponding point cloud in the Point Cloud Layer.

For \textbf{\textit{Segment Layer}}, connectivity of lane graph $\mathcal{M}_{lg}$ forms the foundation for distinguishing segments as $\mathcal{M}_{seg}$. We designate a small section of the path as a subordinate area near inflection and intersection points on the lane graph. Additionally, intersections have been further categorized into "intersections" and "T-intersections". Inflection points are defined as "L-intersections". The stretch of roadway that doesn't intersect with any inflection or intersection points is termed the "straight roadway". The nodes $\textbf{O}$ in the Instance Layer are linked to the closest road segments.

For \textbf{\textit{Environment Layer}}, we construct a global environment node $\mathcal{M}_{env}$ that connects all road segments $\mathcal{M}_{seg}$ in Segment Layer.

Collectively, the aforementioned layers constitute a comprehensive OpenGraph $\mathcal{M}_{all}^{(t)}$ that encompasses a semantic comprehension and a hierarchical topology of the outdoor environment.

\begin{table*}[!t]\normalsize
\scriptsize
\renewcommand{\arraystretch}{1.1}
\centering
\caption{Quantitative Results (IoU and F1 Score) of Semantic Segmentation on SemanticKITTI Dataset}
\label{ss_table}
\resizebox{\textwidth}{!}{\begin{tabular}{p{0.3cm}<{\centering}l<{\centering}p{0.8cm}<{\centering}p{0.8cm}<{\centering}p{0.8cm}<{\centering}p{0.8cm}<{\centering}p{0.8cm}<{\centering}p{0.8cm}<{\centering}p{0.8cm}<{\centering}p{0.8cm}<{\centering}p{0.9cm}<{\centering}p{1.0cm}<{\centering}}
\hline
\textbf{Seq.} & \textbf{Method }
& \cellcolor[RGB]{245, 150, 100} \textcolor{white}{Car}
& \cellcolor[RGB]{255, 0, 255} \textcolor{white}{Road}
& \cellcolor[RGB]{75, 0, 75} \textcolor{white}{Sidewalk}
& \cellcolor[RGB]{0, 200, 255} \textcolor{white}{Building}
& \cellcolor[RGB]{50, 120, 255} \textcolor{white}{Fence}
& \cellcolor[RGB]{0, 175, 0} \textcolor{white}{Veget.}
& \cellcolor[RGB]{150, 240, 255} \textcolor{white}{Pole}
& \cellcolor[RGB]{0, 0, 255} \textcolor{white}{T.-sign}
& \textbf{Average} & \textbf{F1 Score} \\ \hline 
\multirow{6}{*}{03} & RangeNet++ \cite{rangenet++} & 0.4191 & 0.8096 & 0.4656 & 0.3962 & 0.3483 & 0.5410 & 0.1245 & 0.1302 & 0.4780 & 0.6115 \\
 & DeepLab V3 \cite{deeplab} & 0.2535 & 0.6784 & 0.3392 & 0.3262 & 0.0632 & 0.5445 & 0.1270 & 0.1521 & 0.4626 & 0.6025 \\
 & DecoupleSegNets \cite{dsn} & 0.3075 & 0.8283 & 0.5314 & 0.3912 & 0.0646 & 0.6624 & 0.1412 & 0.2457 & 0.6025 & 0.7277 \\
 & CLIP-based \cite{CLIP} & 0.5851 & 0.4349 & 0.0000 & 0.1654 & 0.0823 & 0.3497 & 0.0000 & 0.0138 & 0.3972 & 0.5797 \\
 & BLIP-based \cite{blip} & 0.3215 & 0.4478 & 0.0000 & 0.3240 & 0.0227 & 0.3789 & 0.0122 & 0.1202 & 0.4017 & 0.5673 \\
 & OpenGraph & 0.4191 & 0.7299 & 0.2552 & 0.1737 & 0.6599 & 0.0083 & 0.2960 & 0.0000 & \textbf{0.6051} & \textbf{0.7302} \\ \hline
\multirow{6}{*}{05} & RangeNet++ \cite{rangenet++} & 0.5634 & 0.8575 & 0.5591 & 0.5933 & 0.3605 & 0.5368 & 0.1285 & 0.0950 & 0.4990 & 0.6200 \\
 & DeepLab V3 \cite{deeplab} & 0.3627 & 0.7731 & 0.3749 & 0.5094 & 0.2855 & 0.6066 & 0.0986 & 0.0388 & 0.5203 & 0.6579 \\
 & DecoupleSegNets \cite{dsn} & 0.4489 & 0.8501 & 0.5850 & 0.5228 & 0.3022 & 0.6293 & 0.1260 & 0.0845 & 0.5895 & 0.7212 \\
 & CLIP-based \cite{CLIP} & 0.5618 & 0.0356 & 0.0000 & 0.1769 & 0.3427 & 0.0028 & 0.0312 & 0.0000 & 0.2964 & 0.4518 \\
 & BLIP-based \cite{blip} & 0.5997 & 0.4334 & 0.0000 & 0.3318 & 0.1237 & 0.1926 & 0.0000 & 0.0342 & 0.3320 & 0.4781 \\
 & OpenGraph & 0.7000 & 0.7963 & 0.4823 & 0.7961 & 0.3957 & 0.6312 & 0.1552 & 0.1056 & \textbf{0.6598} & \textbf{0.7749} \\ \hline
\multirow{6}{*}{08} & RangeNet++ \cite{rangenet++} & 0.6339 & 0.8214 & 0.4786 & 0.5500 & 0.1086 & 0.6083 & 0.1607 & 0.1504 & 0.5111 & 0.6295 \\
 & DeepLab V3 \cite{deeplab} & 0.4167 & 0.7162 & 0.3087 & 0.4923 & 0.0736 & 0.6758 & 0.1074 & 0.0691 & 0.5425 & 0.6733 \\
 & DecoupleSegNets \cite{dsn} & 0.4445 & 0.8370 & 0.4870 & 0.5414 & 0.0845 & 0.7152 & 0.1026 & 0.0631 & 0.6376 & 0.7576 \\
 & CLIP-based \cite{CLIP} & 0.3942 & 0.3456 & 0.0000 & 0.2439 & 0.0952 & 0.2166 & 0.0033 & 0.0693 & 0.2753 & 0.4130 \\
 & BLIP-based \cite{blip} & 0.4025 & 0.4318 & 0.0000 & 0.3909 & 0.0633 & 0.2840 & 0.0008 & 0.0422 & 0.3677 & 0.5250 \\
 & OpenGraph & 0.6667 & 0.7436 & 0.3594 & 0.7581 & 0.1447 & 0.6384 & 0.1892 & 0.1635 & \textbf{0.6463} & \textbf{0.7633} \\ \hline
\end{tabular}
}
\end{table*}

\section{Experimental Results}  \label{ER}
In this section, the performance of the proposed OpenGraph is validated through experiments performed on the public ourdoor dataset SemanticKITTI \cite{semantickittiijrr}.
We aim to use experiments to validate our framework, through the following specific questions: 

\begin{enumerate}
  \item Without fine-tuning any models, can OpenGraph accomplish zero-shot 3D semantic understanding?
  \item Does Caption-Enhanced Object Comprehension provide more comprehensible object retrieval? 
  \item What are the potential applications of open-vocabulary hierarchical graph representation?
\end{enumerate}

\subsection{3D Semantic Segmentation}
We first validate OpenGraph's ability for zero-shot semantic understanding in outdoor environments.
As detailed in subsection \ref{hsgf}, we conduct 3D point cloud semantic segmentation at the Point Cloud Layer using 19 predefined classes from the SemanticKITTI dataset. Due to the absence of a direct zero-shot outdoor semantic segmentation comparison baseline, we initially consider two classes of fully supervised methods. The first class comprises a point cloud segmentation technique trained and fine-tuned on the SemanticKITTI dataset, namely \textbf{RangeNet++} \cite{rangenet++}. We generate a global semantic map using its predicted labels. The second class consists of 2D image segmentation methods trained on outdoor image datasets, including \textbf{DeepLab v3} \cite{deeplab} and \textbf{DecoupleSegNets} \cite{dsn}. We derive 3D semantic maps by projecting their image segmentation onto point clouds. Additionally, we compare methods that directly utilize VLM features to replace caption features in OpenGraph, specifically \textbf{CLIP-based} \cite{CLIP} and \textbf{BLIP-based} \cite{blip}.

We selected three sequences from SemanticKITTI: short (03), medium (05), and long (08). As shown in Fig. \ref{ss}, due to point-wise segmentation in 2D or 3D, the fully supervised approaches generate many scattered points. Additionally, small objects like bicycles suffer from lower segmentation accuracy due to limited training samples. While CLIP-based and BLIP-based, two methods employing VLM features for scene understanding, fail to achieve accurate object classification, OpenGraph excels in this aspect with the assistance of LLM features. 
Tab. \ref{ss_table} presents the results of semantic segmentation quantification, including IoU for common classes and overall F1 scores. Notably, OpenGraph outperforms even comparable fully supervised methods across these sequences. These results demonstrate that OpenGraph achieves accurate zero-shot semantic understanding in outdoor environments.

\begin{figure*}[!t]\centering
	\includegraphics[width=17cm]{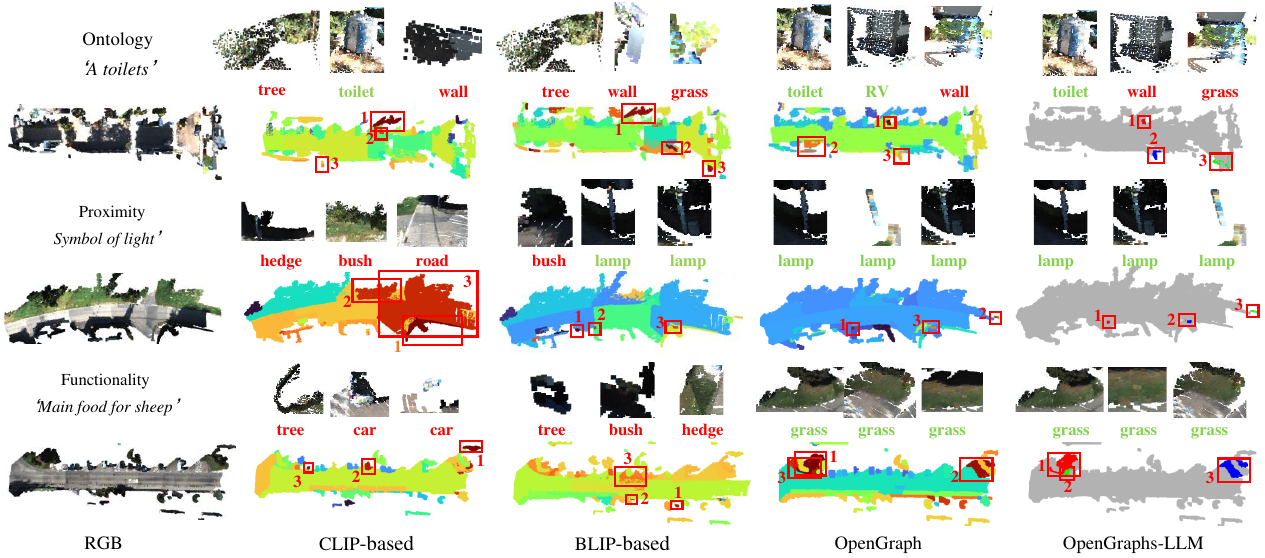}
	\caption{The outcomes from various open-vocabulary text queries (displaying the Top-3 objects). In the visual representation, OpenGraph-LLM highlights the Top-3 objects in red, green, and blue, while the other methods render all objects based on relevance. 
    The text beneath each retrieved object in the figure corresponds to its actual category, where green signifies successful retrieval and red indicates failure.
} 
	\label{or} 
\end{figure*}

\begin{table}[]
\centering
\caption{Quantitative Results (top-1,2,3 recall) of Object Retrieval}
\label{or_table}
\renewcommand\arraystretch{1.1}
\setlength{\tabcolsep}{0.9mm}{
\begin{tabular}{cclcccc}
\hline
\multicolumn{1}{l}{\textbf{Seq.}} & \multicolumn{1}{l}{\textbf{Query-Type}}& \textbf{Methods}& \multicolumn{1}{l}{\textbf{R@1}} & \multicolumn{1}{l}{\textbf{R@2}} & \multicolumn{1}{l}{\textbf{R@3}} & \multicolumn{1}{l}{\textbf{\#Query}} \\ \hline
\multirow{12}{*}{03/05/08}& \multirow{4}{*}{Onotology}& CLIP-based\cite{CLIP}& 0.60 & 0.70 & 0.73 & \multirow{4}{*}{30}\\
&& BLIP-based\cite{blip}& 0.50& 0.60& 0.70&\\
 && OpenGraph& \textbf{0.90}& \textbf{0.90}& \textbf{0.90}&\\
 && OpenGraph-LLM & 0.70& 0.80& 0.87&\\ \cline{2-7} 
 & \multirow{4}{*}{Proximity}& CLIP-based\cite{CLIP}& 0.37& 0.47& 0.53& \multirow{4}{*}{30}\\
 && BLIP-based\cite{blip}& 0.30& 0.40& 0.47&\\
 && OpenGraph& \textbf{0.80}& \textbf{0.87}& \textbf{0.87}&\\
 && OpenGraph-LLM & 0.70& 0.73& 0.80&\\ \cline{2-7} 
 & \multicolumn{1}{l}{\multirow{4}{*}{Functionality}} & CLIP-based\cite{CLIP}& 0.47& 0.53& 0.57& \multirow{4}{*}{30}\\
 & \multicolumn{1}{l}{} & BLIP-based\cite{blip}& 0.37& 0.37& 0.47&\\
 & \multicolumn{1}{l}{} & OpenGraph& 0.63& \textbf{0.80}& 0.80&\\
 & \multicolumn{1}{l}{} & OpenGraph-LLM & \textbf{0.67}& \textbf{0.80}& \textbf{0.87}&\\ \hline
\end{tabular}
}
\end{table}

\subsection{Open-vocabulary Object Retrieval}
To illustrate the advantages of OpenGraph for complex semantic queries, we conducted object retrieval experiments that center around three distinct text types:

\textbf{Ontology:} A direct description of the object itself. For instance, \textit{A tall tree}.

\textbf{Proximity:} The description of the object relies on objects related to it. For example, \textit{I want to find some green leaves} (a tree or a bush).

\textbf{Functionality:} The description focuses on the inferred functions of the object in question. For instance, \textit{Something that can be used for driving} (a car).

The experiments involve small segments extracted from the three sequences, each containing a diverse array of objects.  For each query type on each sequence, we generated 10 distinct descriptions, with relevant objects manually chosen as ground truth values. The Fig. \ref{or} illustrates some of the comparison results of OpenGraph with other zero-shot methods, where the colors are rendered according to the degree of similarity of the features (VLM features for CLIP-based and BLIP-based, LLM features for OpenGraph). 
Since OpenGraph maintains an explicit caption for each object, it facilitates effortless access to LLM for object retrieval, denoted as \textit{OpenGraph-LLM}. To elaborate, we feed both the map object captions and query text into GPT4 \cite{GPT4}, tasking it with identifying the top three most pertinent objects, labeled red, green, and blue, respectively.
Tab. \ref{or_table} presents the overall top-1, top-2, and top-3 recall measurements.

\begin{figure}[!t]\centering
	\includegraphics[width=8.0cm]{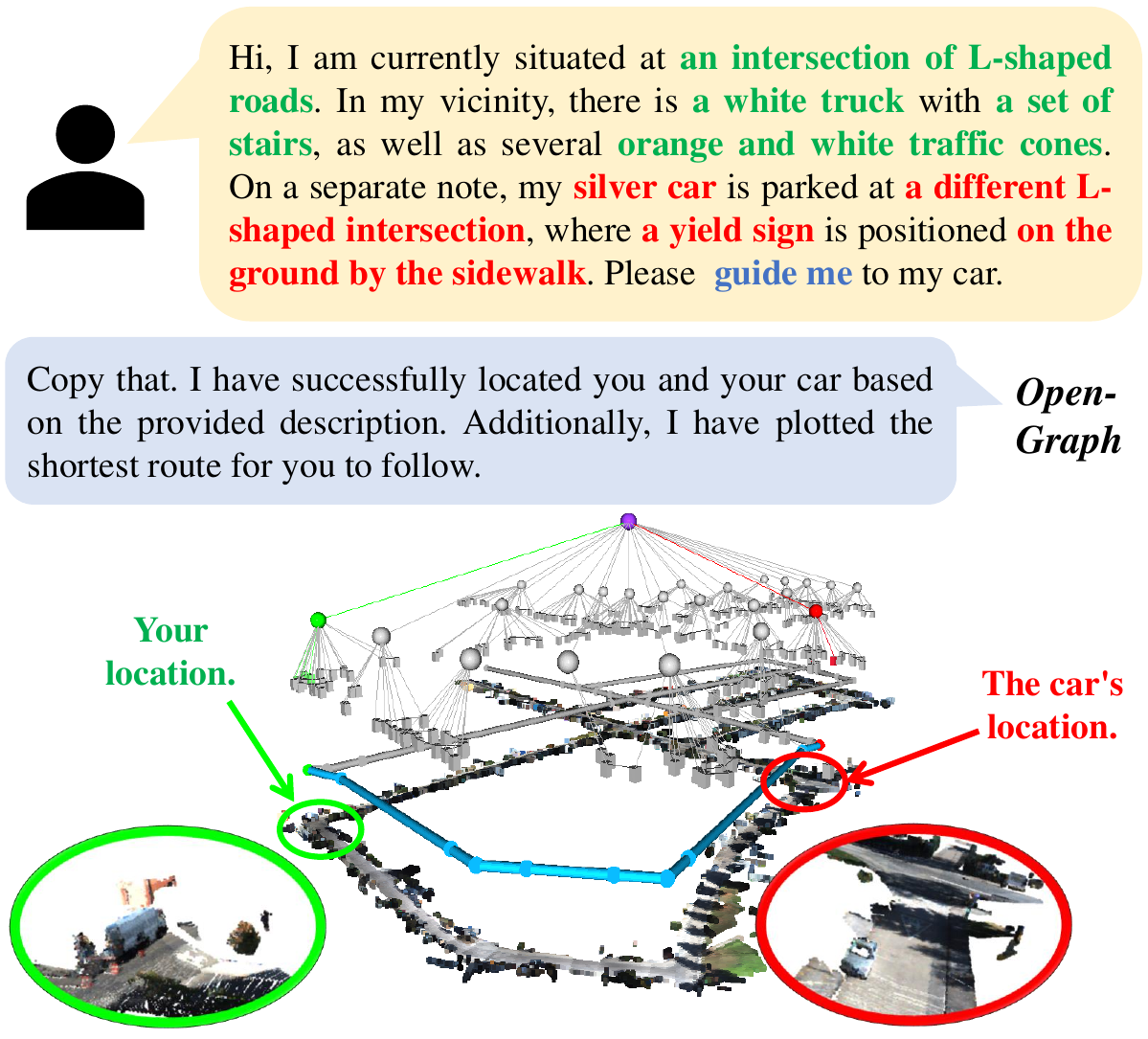}
	\caption{OpenGraph efficiently identifies the start  (green) and end  (red) points by utilizing user descriptions and performs path planning to determine the optimal route (blue) between them.} 
\label{pp}    
\end{figure}

First, in ontology queries, OpenGraph demonstrates the highest recall rate, mitigating the positional bias when LLM is applied, such as recognizing a toilet in a previously observed RV. Secondly, in proximity queries, the performance of VLM-based methods sharply drops, due to limited natural language comprehension like 'symbol of light' versus 'lamp'. Lastly, OpenGraph excels in challenging functional queries with LLM boosting its natural language reasoning. In summary, OpenGraph and its LLM variants demonstrate superior natural language reasoning in a variety of outdoor object retrieval tasks.

\begin{figure}[!t]\centering
	\includegraphics[width=7.8cm]{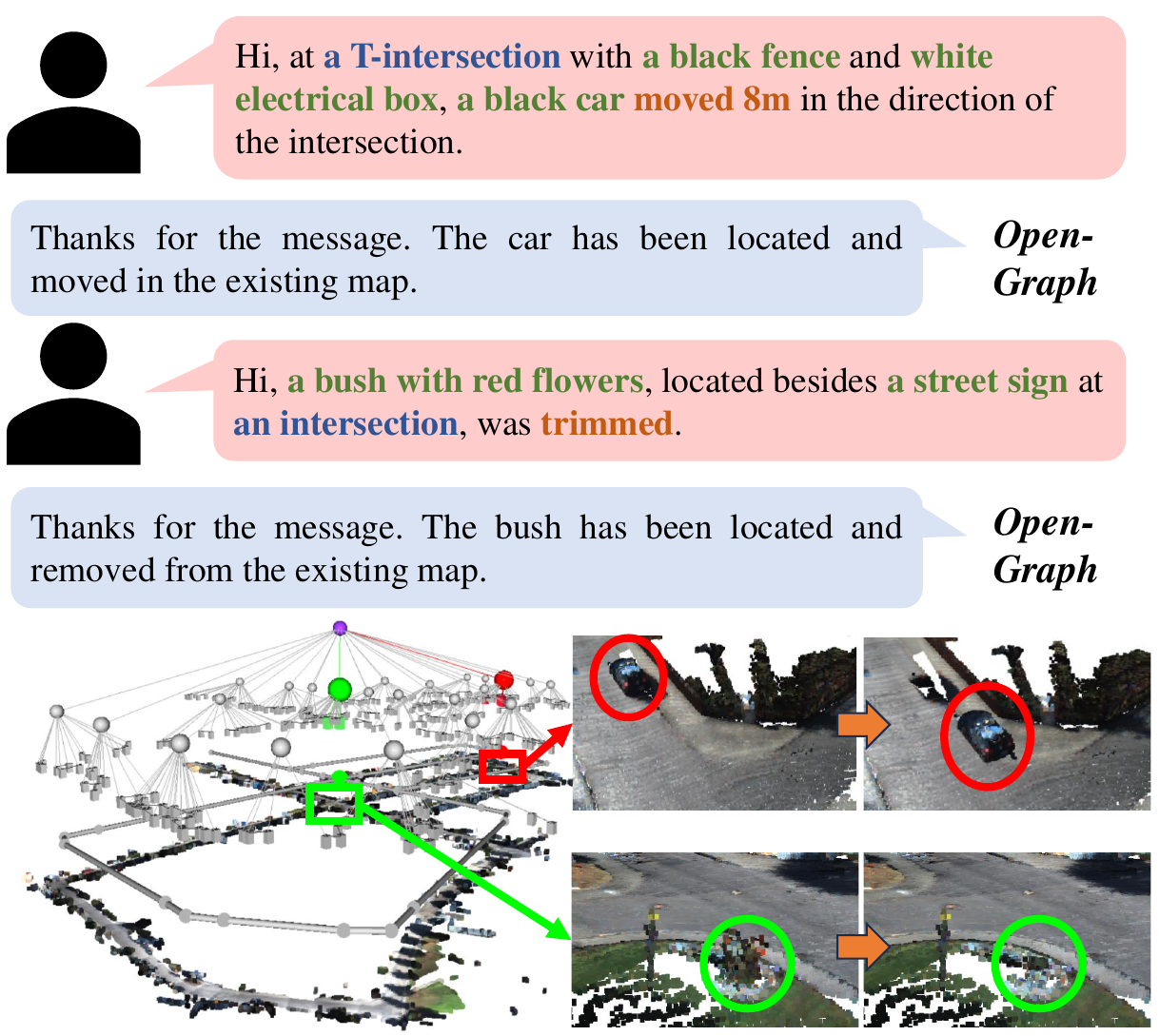}
	\caption{OpenGraph's open-vocabulary hierarchical map representation facilitates human-interactive map updating. Here shows the updated point cloud in the Point Cloud Layer.} 
\label{mu}    
\end{figure}

\subsection{Hierarchical Graph Structured Query}
Hierarchical graphs offer the benefit of structured top-down queries. By integrating with LLMs on the front end, OpenGraph empowers users to accomplish a wide range of tasks more effectively and efficiently. Imagine the queriers situated within a scene, capable of observing only a limited portion of their surroundings. By indicating the type of road (Segment Layer) they are on and the relationships between objects nearby (Instance Layer), they can swiftly localize themselves. Moreover, if they can provide hints about their destination, OpenGraph can facilitate global path planning on the lane graph to guide them effectively. A related case is shown in Fig. \ref{pp}.

Furthermore, OpenGraph facilitates human-interactive map updating. Individuals within the environment can contribute the latest map patches to OpenGraph by detecting changes in their surroundings, ensuring the continual upkeep of the environment map. Fig. \ref{mu} visually demonstrates this process, where a person actively observes the removal of a bush beneath a street sign at an intersection. OpenGraph promptly identifies the object based on the description and promptly updates the map accordingly.

\section{Conclusion} \label{C}

This paper introduces OpenGraph, an open vocabulary hierarchical 3D graph representation framework for large-scale outdoor environments. Initially, the extraction of instance masks, captions, and features occurs in the 2D images, facilitated by VLMs and LLMs. Subsequently, an object-centric map is incrementally constructed and fused by projecting onto 3D point clouds. Finally, the extraction of the lane graph and subsequent scene segmentation culminates in the derivation of a hierarchical graph. The results from evaluations on public datasets reveal that OpenGraph, operating as a zero-shot method, even exhibits superior performance in 3D semantic segmentation compared to fully supervised methods, while also demonstrating advantages in open-vocabulary retrieval over alternative types. Moreover, the hierarchical graph representation facilitates rapid structured queries.
Moving forward, we need to inject richer semantics at the edges between instances and provide component-level understanding.


\bibliographystyle{Bibliography/IEEEtran}
\bibliography{Bibliography/RAL}

\end{document}